\documentclass[runningheads]{llncs}

 \usepackage{eccv}

\usepackage{eccvabbrv}

\usepackage{graphicx}
\usepackage{booktabs}

\usepackage[accsupp]{axessibility}  

\usepackage[pagebackref]{hyperref}

\usepackage{orcidlink}

\begin{document}

\title{PseudoMapLabeler: Confidence-Aware Pseudo-Label Generation for Semi-Supervised Online Mapping} 

\titlerunning{PseudoMapLabeler}

\author{Chikao Tsuchiya\inst{1}\orcidlink{0000-1111-2222-3333} \and
Dhaval Bhanderi\inst{1}\orcidlink{1111-2222-3333-4444} \and
David Ilstrup\inst{1}\orcidlink{2222--3333-4444-5555} \and
Hsinmin Cheng\inst{1}\orcidlink{2222--3333-4444-5555} \and 
Christopher Ostafew\inst{1}\orcidlink{2222--3333-4444-5555}}

\authorrunning{C. Tsuchiya et al.}

\institute{Nissan Advanced Technology Center - Silicon Valley, Nissan North America, Santa Clara, CA 95014, USA \and
\email{\{Chikao.Tsuchiya, Dhaval.Bhanderi, David.Ilstrup, Hsinmin.Cheng, Christopher.Ostafew\}@nissan-usa.com}\\
}

\maketitle

\begin{abstract}
A critical challenge in deploying online HD map construction systems to real-world scenarios is the scarcity of labeled training data, which limits model generalization in diverse environments. To address this limitation, we propose a teacher-student semi-supervised learning (SSL) framework that generates high-quality pseudo-labels from unlabeled data through confidence-aware map refinement. Our approach first trains a teacher model on limited labeled data, then leverages Beta-distribution-based confidence maps to assess the reliability of predicted map elements across temporal observations. Unlike conventional filtering methods that discard entire elements, we introduce a spatial clipping technique that selectively preserves high-confidence regions while removing unreliable segments. The refined map elements serve as map priors that improve the teacher model's prediction accuracy on unlabeled data in a second pass. These enhanced predictions become pseudo-labels for training a student model from scratch, followed by fine-tuning on the original labeled data. Experimental results on the nuScenes dataset demonstrate that our teacher-student framework with refined pseudo-labels improves performance by +6.1 mAP under a low-label regime compared to training on labeled data alone, offering a practical solution to the labeled data scarcity problem in online HD map construction.
\end{abstract}

\section{Introduction}
\label{sec:intro}

Online HD map construction has emerged as a critical component for autonomous driving systems, enabling real-time perception of road topology and semantic elements from onboard sensors~\cite{elghazaly2023high}. Recent advances in deep learning-based approaches have demonstrated impressive performance on benchmark datasets such as nuScenes~\cite{caesar2020nuscenes} and Argoverse~\cite{wilson2023argoverse}. However, a fundamental challenge remains when deploying these systems to real-world scenarios: the scarcity of labeled training data. Annotating HD maps requires extensive manual effort by domain experts, making it prohibitively expensive to collect labeled data for every new environment or geographic region. This data scarcity severely limits the generalization capability of trained models, hindering their practical deployment.

Semi-supervised learning offers a promising solution by leveraging abundant unlabeled data alongside limited labeled samples. The key challenge lies in generating high-quality pseudo-labels from model predictions on unlabeled data. Naive approaches that directly use raw predictions as pseudo-labels often introduce significant noise, leading to error accumulation during training~\cite{scherer2022pseudo, arazo2020pseudo, sun2025semi}. For online HD map construction, this problem is particularly acute due to the temporal nature of predictions: while individual frame predictions may contain errors, aggregating predictions across multiple temporal observations can provide more reliable estimates. However, simple temporal aggregation without confidence assessment fails to distinguish between consistent high-quality predictions and noisy outliers.

In this work, we propose PseudoMapLabeler (PML), a teacher-student semi-supervised learning framework that generates high-quality pseudo-labels through confidence-aware map prior refinement for online HD map construction. Unlike conventional approaches that directly use model predictions as pseudo-labels, PML first constructs refined map priors from temporally aggregated predictions and leverages these priors to guide the teacher model in generating more accurate pseudo-labels. Specifically, PML temporally aggregates predictions from an initial teacher model trained on limited labeled data, constructs Beta-distribution-based confidence maps to assess spatial reliability, and applies spatial clipping to extract high-confidence map segments as refined priors. These refined priors are then fed back to the teacher model to improve its predictions on unlabeled data, which serve as pseudo-labels for training a student model. This two-stage pipeline—map prior refinement followed by pseudo-label generation—enables effective utilization of unlabeled data while maintaining prediction quality. We conduct extensive experiments on the nuScenes dataset to validate the effectiveness of our approach, demonstrating significant improvements over baseline methods in semi-supervised settings. Our key contributions are as follows:

\textbf{1. Beta-Distribution Confidence Map:} We propose a Beta-distribution based evidence aggregation scheme that summarizes spatiotemporal confidence over a BEV grid and yields a per-cell reliability score for map refinement. Unlike deterministic approaches, our method accounts for both the number of observations and the prediction confidence at each spatial location, providing an interpretable way to assess reliability.

\textbf{2. Spatial Clipping for Prior Map Generation:} We propose a novel spatial clipping technique that selectively preserves high-confidence regions of map elements while removing unreliable segments. To handle class-specific confidence distributions, we employ percentile-based adaptive thresholding that ensures balanced prior map generation across different map element types. This approach maximizes the utilization of unlabeled data by retaining partial information from predictions, rather than discarding entire elements as in conventional filtering methods.

\textbf{3. Model-Agnostic Pseudo-Labeling Pipeline:} Our pseudo-label generation and refinement procedure is architecture-agnostic: it operates on predicted vectorized map elements and does not rely on model-specific internals. While we instantiate the teacher–student framework with Uni-PrevPredMap ~\cite{peng2025uni} in our experiments for controlled evaluation, the refined pseudo-labels can, in principle, be used to train other online vectorized mapping architectures. We additionally validate this on a second architecture (MapTR~\cite{liao2023maptr}), observing consistent gains.

\section{Related Work}

\subsection{Online HD Map Construction}
Online HD map construction aims to predict vectorized map elements directly from onboard sensor observations without relying on pre-built maps. Early approaches such as HDMapNet~\cite{li2022hdmapnet} and VectorMapNet~\cite{liu2023vectormapnet} pioneered end-to-end learning frameworks that output map elements in bird's-eye view (BEV) representation. Transformer-based designs further improved accuracy and efficiency. MapTR~\cite{liao2023maptr} proposed permutation-equivalent modeling and hierarchical query embeddings for robust vectorized extraction. MapTRv2~\cite{liao2025maptrv2} added auxiliary one-to-many matching and dense supervision to accelerate convergence and raise SOTA on nuScenes~\cite{caesar2020nuscenes} and Argoverse2~\cite{wilson2023argoverse}. These methods typically process single frames independently, limiting their ability to leverage temporal information for improved accuracy and consistency.

Recent works emphasize temporal stability and long-range coverage. StreamMapNet~\cite{yuan2024streammapnet} performs streaming temporal modeling with multi-point attention, and also highlighted the importance of geographically disjoint evaluation splits. MapTracker~\cite{chen2024maptracker} treats mapping as tracking with strided memory fusion over BEV and vector latents, yielding consistent reconstructions across frames. To inject priors, Neural Map Prior (NMP)~\cite{xiong2023neural} maintains a learnable global prior that is queried during online inference and updated by a learning-based fusion module and retrieved via cross-attention. Complementary to that, P-MapNet~\cite{jiang2024p} leverages both low-cost SD maps and an HD-map prior learned via masked autoencoding to improve far-range map quality. Beyond geometric priors, ~\cite{tumu2025usinglanguageroadmanuals}
adds linguistic priors -- structured OSM road metadata and
lane-width conventions from road design manuals. These trends indicate a shift from single-frame, purely local inference toward temporal- and prior-aware online mapping.

\subsection{Temporal Modeling and Prior Fusion for Online Mapping}
Temporal cues are crucial for mitigating occlusions and improving structural consistency. In perception, BEVFormer~\cite{bevformer} introduced spatiotemporal transformers that recurrently fuse historical BEV features. PETRv2~\cite{liu2023petrv2} generalized temporal alignment using 3D positional encodings and provided a unified interface to BEV segmentation and 3D lane detection. For vectorized mapping specifically, PrevPredMap~\cite{peng2025prevpredmap} proposed to use previous predictions themselves as high-level priors, with a previous-prediction-based query generator and a dynamic-position-query decoder. Extending this idea, Uni-PrevPredMap ~\cite{peng2025uni} unified temporal predictions and map priors under a tri-mode design, together with a tile-indexed global processor for efficient prior refresh and retrieval. D2HDMap~\cite{d2hdmap} injects a lightweight driveline-only map prior alongside the temporal prior to guide vectorized map prediction. HRMapNet~\cite{zhang2024enhancing} conditions on historical outputs stored explicitly as rasterized priors to initialize queries and enhance features. Our teacher-side ``prior'' belongs to this temporal/prior fusion family but differs in that the priors are self-refined offline via confidence-aware temporal accumulation before being re-injected to guide online predictions.

\subsection{Semi-Supervised Learning for Online Mapping}
Label scarcity motivates semi-supervised and weakly-supervised approaches. Semi-supervised learning has been widely studied in computer vision tasks such as image classification and object detection~\cite{kage2024review,arazo2020pseudo,pseudolabel,meanteacher}, but its application to online HD map construction remains largely unexplored~\cite{lilja2025exploring,lowens2025pseudomaptrainer}. The primary challenge lies in generating reliable pseudo-labels for vectorized map elements, which have complex geometric structures unlike simple class labels or bounding boxes.

Recent works have begun to address this challenge. The Teacher-Student paradigm forms one approach to semi-supervised learning (SSL) for online mapping, where a student model learns from pseudo-labels generated by a teacher model~\cite{lilja2025exploring}. Strong augmentation strategies such as CamDrop~\cite{ishikawa2024pct}, CutOut~\cite{devries2017improved}, and BEVDrop~\cite{yang2023revisiting} are employed to increase task difficulty while maintaining geometric consistency. 
PseudoMapTrainer~\cite{lowens2025pseudomaptrainer} leverages Gaussian Splatting~\cite{mei2024rome} to reconstruct road surfaces and render semantic BEV maps for automatic vectorization. It models partial map element observation via a mask-aware loss to assemble complete map elements from partial elements during temporal aggregation. 
Vectorized Map Annotation (VMA) framework~\cite{chen2023vma} achieves SSL via a human-in-the-loop verification scheme to minimize manual annotation. In contrast to ~\cite{lilja2025exploring}, which employs an EMA-based online teacher operating on rasterized BEV segmentation output, our method uses a fixed offline teacher on vectorized polyline representations, enabling segment-level pseudo-label refinement via explicit Beta-distribution confidence modeling rather than the multi-frame confidence fusion over rasterized grids. These architectural differences preclude direct numerical comparison. Nevertheless, both works demonstrate the viability of SSL for online mapping under label scarcity.

\section{Method}

\begin{figure*}[t]
\centering
\includegraphics[width=0.95\textwidth]{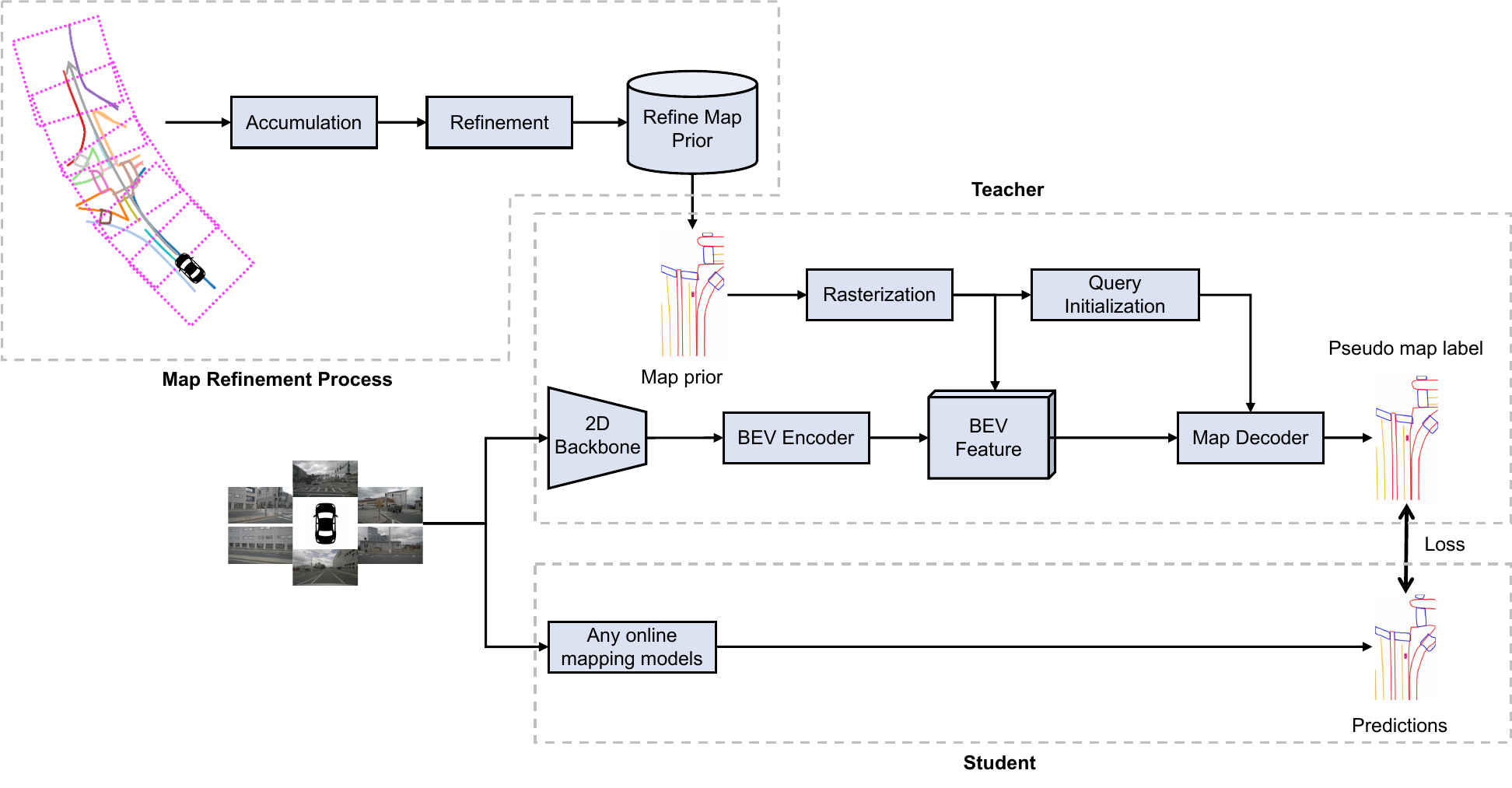}
\caption{Overview of the proposed teacher-student semi-supervised learning framework. The teacher model is based on Uni-PrevPredMap (UPPM)~\cite{peng2025uni}, which incorporates map prior rasterization and prior-based query initialization mechanisms. However, instead of relying on ground truth HD map priors, we equip the teacher with refined map priors generated through our Map Refinement Process. An initial teacher is first trained on a small amount of labeled data, and the refined map priors are then constructed by accumulating the initial teacher's predictions on unlabeled data and applying confidence-based spatial clipping. Since the teacher produces vectorized map labels as pseudo-labels, any online vectorized mapping model can be employed as the student.}
\label{fig:overview}
\end{figure*}

\subsection{Problem Formulation}

We formulate the problem of confidence-aware pseudo-label generation for semi-supervised online HD map construction using a teacher-student framework. We assume access to a limited labeled dataset $\mathcal{D}_L = \{(\mathcal{F}_i, \mathcal{M}_i^{gt})\}_{i=1}^{N_L}$ and a larger unlabeled dataset $\mathcal{D}_U = \{\mathcal{F}_j\}_{j=1}^{N_U}$, where $N_U \gg N_L$. Each sequence $\mathcal{F}$ consists of frames $\{F_1, F_2, \ldots, F_T\}$ with corresponding ego-vehicle poses $\mathcal{P} = \{(t_1, R_1), \ldots, (t_T, R_T)\}$, where $t_i \in \mathbb{R}^3$ and $R_i \in SO(3)$ represent translation and rotation respectively. The ground truth map elements $\mathcal{M}^{gt}$ consist of vectorized polylines with semantic class labels $c \in \{\text{divider}, \text{ped\_crossing}, \text{boundary}\}$.

Our goal is to generate high-quality pseudo-labels $\hat{\mathcal{M}}$ from unlabeled data $\mathcal{D}_U$ that can be used to train a student model, ultimately improving performance beyond what is achievable with labeled data alone.

\subsection{Method Overview}

Fig.~\ref{fig:overview} illustrates the overall pipeline of our teacher–student framework for semi-supervised online mapping, which consists of the following steps.

\begin{enumerate}
\item \textbf{Teacher Model Training:} Train a teacher model on $\mathcal{D}_L$ and calibrate its confidence scores.

\item \textbf{Map Prior Generation:} Apply the trained teacher model on $\mathcal{D}_U$ and accumulate predictions. Then, generate map priors via confidence-aware spatial clipping.

\item \textbf{Pseudo-Label Generation:} Re-apply the teacher model to $\mathcal{D}_U$ with the refined map priors to generate pseudo-labels $\hat{\mathcal{D}}_U$.

\item \textbf{Student Model Training:} Pre-train a student model on $\hat{\mathcal{D}}_U$, then fine-tune it on $\mathcal{D}_L$.
\end{enumerate}

While we instantiate both teacher and student with UPPM in our experiments for controlled comparison, the proposed refinement-and-pseudo-labeling pipeline is plug-and-play for other vectorized online mapping architectures. The following sections describe the details of each component.

\subsection{Teacher Model}

Our teacher model builds upon PrevPredMap (PPM)~\cite{peng2025prevpredmap} and its unified extension Uni-PrevPredMap (UPPM)~\cite{peng2025uni}, which are state-of-the-art online vectorized mapping methods that leverage temporal priors. These methods can optionally incorporate prior map information $\mathcal{M}_{prior}$ to improve prediction accuracy by providing contextual guidance about expected map element locations.

The model predicts map elements $\mathcal{M}_i = \{m_i^1, m_i^2, \ldots, m_i^{N_i}\}$ for each frame $F_i$, where each element $m_i^j$ consists of a polyline represented by a sequence of 2D points $\{p_1, p_2, \ldots, p_K\}$ in the ego-vehicle coordinate frame, along with a semantic class label and a prediction confidence score $s \in [0, 1]$.

In our framework, we exploit this prior-conditioning capability in a novel way: rather than using ground truth HD maps as priors (which are unavailable for unlabeled data), we generate \textit{refined map priors} from the model's own predictions through temporal aggregation and confidence-based filtering. This self-refinement mechanism enables the teacher model to progressively improve its predictions on unlabeled data without requiring additional supervision.

\subsection{Temporal Accumulation}

To aggregate map elements across frames, we transform predictions from ego-vehicle coordinates to a common reference frame. Importantly, our method does not require a globally consistent coordinate system spanning large geographic areas. Instead, we operate within local scenes that cover limited spatial extents, typically corresponding to a single city block or intersection.

For the nuScenes dataset, sequences are divided into 20-second scenes, which we adopt as our processing unit. Within each scene, we transform map elements from frame $i$ to a scene-local coordinate system using the provided ego-vehicle poses:
\begin{equation}
p_{scene} = R_i \cdot p_{ego} + t_i
\end{equation}
where $p_{ego}$ is a point in ego-vehicle coordinates and $p_{scene}$ is the corresponding point in the scene-local coordinate system. We denote the transformed map elements from frame $i$ as $\mathcal{M}_i^{scene}$. The accumulated map elements $\mathcal{M}_{acc} = \bigcup_{i=1}^{n_{frm}(scene)} \mathcal{M}_i^{scene}$ form the input to our confidence estimation module.

\subsubsection{Probability Calibration}

Neural network outputs often produce overconfident or poorly calibrated probability estimates. To address this issue, we apply temperature scaling~\cite{guo2017calibration}, a simple yet effective post-hoc calibration method. Given the raw logit output $z_m$ from the network, we apply a temperature parameter $T > 0$ before the sigmoid activation:
\begin{equation}
s_m^{calib} = \sigma(z_m / T) = \frac{1}{1 + \exp(-z_m / T)}
\end{equation}
where $\sigma(\cdot)$ denotes the sigmoid function. The temperature parameter $T$ is optimized on a held-out validation set by minimizing the negative log-likelihood of binary cross-entropy (NLL-BCE):
\begin{equation}
\mathcal{L}_{calib}(T) = -\frac{1}{N} \sum_{i=1}^{N} \left[ y_i \log(s_i^{calib}) + (1-y_i) \log(1-s_i^{calib}) \right]
\end{equation}
where $y_i \in \{0, 1\}$ denotes the ground truth label and $N$ is the number of validation samples. In our experiments, we use the nuScenes validation set for this optimization.


\begin{figure*}[t]
\centering
\includegraphics[width=0.8\textwidth]{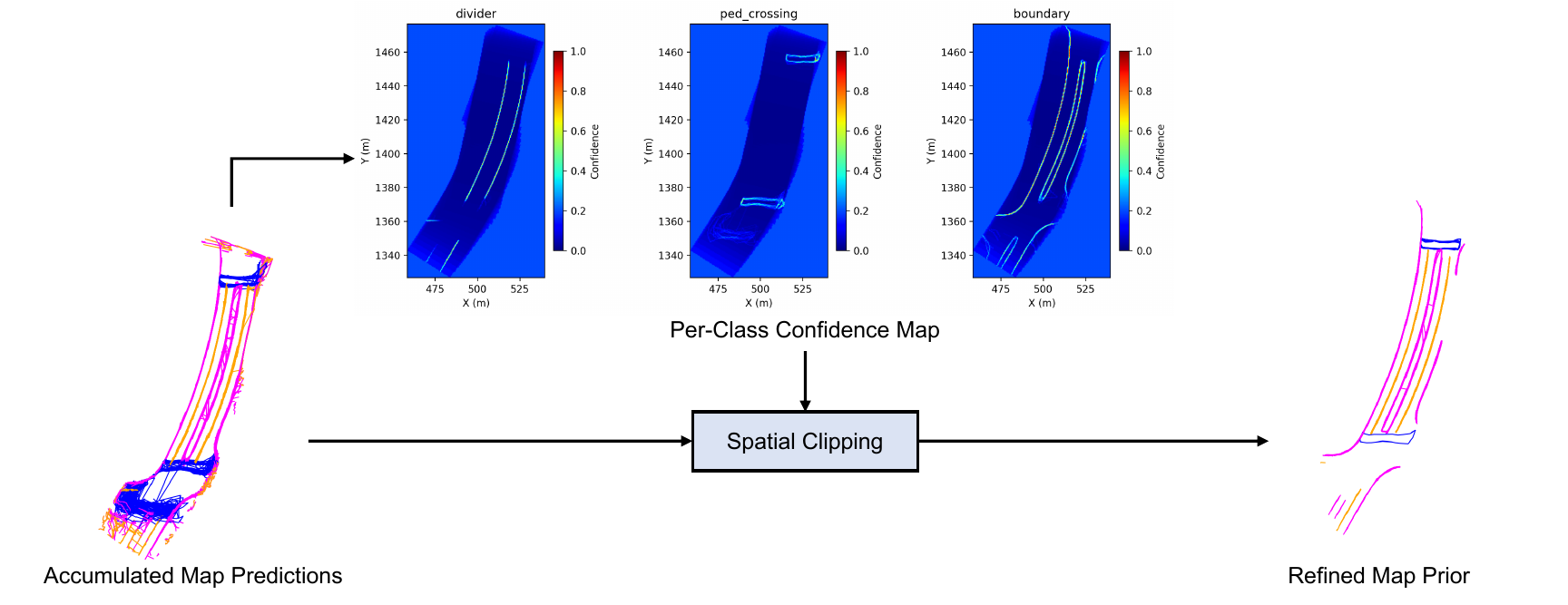}
\caption{\textbf{Visualization of the confidence-aware map refinement process.} Left: temporally accumulated map predictions across frames exhibit noise and duplication.
Top: per-class Beta-distribution-based confidence maps estimated from temporal observations.
Right: refined map elements obtained via spatial clipping, preserving high-confidence segments while removing unreliable regions.}
\label{fig:map_refinementt}
\end{figure*}

\subsection{Beta-Distribution Confidence Map}

Given $\mathcal{M}_{acc}$, we construct a per-class spatial confidence map over a regular grid (resolution $\delta$ = 0.5m). For each grid cell $(x, y)$ and class $c$, we track two quantities: the number of times the cell was observed (i.e., within the configured BEV spatial extent for that frame) $n_{obs}(x, y, c)$ and the weighted detection count based on calibrated prediction confidence scores.

We model the detection probability at each grid cell using a Beta distribution, which naturally represents uncertainty in binomial processes. The Beta distribution is parameterized by shape parameters $\alpha$ and $\beta$, which we update based on observations:

\begin{align}
\alpha(x, y, c) &= \alpha_0 + \sum_{i=1}^{n_{frm}(scene)} \sum_{m \in \mathcal{M}_i^{scene}} s_m^{calib} \cdot \mathbf{1}_{(x,y) \in m, c_m = c} \\
\beta(x, y, c) &= \beta_0 + n_{obs}(x, y, c) - (\alpha(x, y, c) - \alpha_0)
\end{align}

where $\alpha_0$ and $\beta_0$ are prior parameters that encode our initial belief about detection probability. We set $\alpha_0 = p_0 \cdot \kappa$ and $\beta_0 = (1 - p_0) \cdot \kappa$, where $p_0$ is the prior detection probability and $\kappa$ is the prior strength. In our implementation, we use $p_0 = 0.2$ and $\kappa = 2.0$, reflecting a weak prior biased toward absence.

The confidence at each grid cell is then computed as the posterior mean of the Beta distribution:
\begin{equation}
\text{conf}(x, y, c) = \frac{\alpha(x, y, c)}{\alpha(x, y, c) + \beta(x, y, c)}
\end{equation}

This formulation has several desirable properties: (1) it naturally accounts for both the frequency of detections and the prediction confidence scores, (2) it provides higher confidence when multiple consistent observations are available, (3) it gracefully handles sparse observations through the Bayesian prior, and (4) it produces confidence estimates that reflect detection frequency and prediction scores.

\subsection{Confidence-Based Map Refinement}

Given the $\text{conf}(x, y, c)$, we propose spatial clipping at the segment level (Fig. 2) to refine accumulated map elements into pseudo-labels, and compare it against element-level filtering.

\subsubsection{Spatial Clipping (Proposed)}

Our spatial clipping method selectively preserves high-confidence portions of map elements while discarding unreliable segments. For each accumulated map element $m \in \mathcal{M}_{acc}$ represented as a polyline $\{p_1, p_2, \ldots, p_K\}$, we sample the confidence value at each point using bilinear interpolation:
\begin{equation}
c_i = \text{BilinearSample}(\text{conf}(\cdot, \cdot, c_m), p_i)
\end{equation}
where $c_m$ is the semantic class of element $m$. We then identify continuous segments where confidence exceeds a threshold $\tau_c$. Each segment defines a contiguous subsequence of points that forms a clipped map element. We apply additional filtering to remove segments that are too short (length $< l_{min}$) or have too few points (count $< n_{min}$).

For polygon-type elements (e.g., pedestrian crossings), we adopt a stricter criterion: we retain the polygon only if all points satisfy $c_i \geq \tau_c$. This all-or-nothing approach is motivated by the semantic nature of polygons, where partial regions do not constitute valid map elements.

\textit{Percentile-Based Thresholding:} A critical challenge in setting the confidence threshold $\tau_c$ is that different map element classes exhibit different confidence distributions. To address this, we use a percentile-based adaptive thresholding strategy. For each class $c$, we collect all confidence values sampled from accumulated map elements of that class, and set the threshold as:
\begin{equation}
\tau_c = \text{Percentile}(\{c_i\}_{c}, 100 - p)
\end{equation}
where $p \in [0, 100]$ is the desired retention percentage. For example, $p = 10$ retains the top 10\% of points for each class.

\subsubsection{Element Filtering (Baseline)}

As a baseline, we compare against an element-level filtering approach that makes binary decisions for entire map elements. For each accumulated element $m$, we compute an aggregated confidence score by sampling confidence values at all points along the polyline and applying a mean function $\mathcal{A}$:
\begin{equation}
\bar{c}_m = \text{Mean}(\{c_1, c_2, \ldots, c_K\})
\end{equation}
An element is retained if $\bar{c}_m \geq \tau_c$, and discarded otherwise. While this approach is simpler to implement, it cannot distinguish between partially reliable and entirely unreliable elements.

\subsection{Teacher-Student Framework}

We now describe how the confidence-aware map refinement components are integrated into a complete teacher-student framework for semi-supervised learning.

\textbf{Teacher Model Training:} We train a teacher model $f_\theta^{teacher}$ using only the labeled dataset $\mathcal{D}_L$. The teacher model is based on UPPM~\cite{peng2025uni}, but instead of using ground truth HD map priors, it will be equipped with refined map priors generated through our PML pipeline. Initially, the teacher is trained from scratch without any map priors (external map priors are injected only during the pseudo-label generation pass on unlabeled data, Step 3). 

\textbf{Map Prior Generation:} We apply the trained teacher model to the unlabeled dataset $\mathcal{D}_U$ to obtain initial predictions:
\begin{equation}
\mathcal{M}_j^{(1)} = f_\theta^{teacher}(\mathcal{F}_j), \quad \forall \mathcal{F}_j \in \mathcal{D}_U
\end{equation}
These predictions are calibrated using the temperature parameter $T^*$. We then aggregate predictions across temporal sequences within each scene to form accumulated map elements $\mathcal{M}_{acc}$. From the accumulated predictions, we construct Beta-distribution-based confidence maps for each scene and semantic class. We then apply spatial clipping with percentile-based thresholding to extract high-confidence map element segments:
\begin{equation}
\mathcal{M}_{prior} = \text{SpatialClip}(\mathcal{M}_{acc}, \text{conf}(\cdot, \cdot, \cdot), p)
\end{equation}
The resulting refined map elements $\mathcal{M}_{prior}$ are converted to a tile-based format suitable for use as temporal priors in UPPM.

\textbf{Pseudo-Label Generation:} We apply the teacher model again to the unlabeled dataset, but this time providing the refined temporal priors $\mathcal{M}_{prior}$ as additional input:
\begin{equation}
\mathcal{M}_j^{(2)} = f_\theta^{teacher}(\mathcal{F}_j, \mathcal{M}_{prior}), \quad \forall \mathcal{F}_j \in \mathcal{D}_U
\end{equation}
The temporal priors guide the model's attention toward regions where reliable map elements have been consistently detected, leading to improved prediction accuracy. These enhanced predictions serve as pseudo-labels:
\begin{equation}
\hat{\mathcal{D}}_U = \{(\mathcal{F}_j, \mathcal{M}_j^{(2)})\}_{j=1}^{N_U}
\end{equation}

\textbf{Student Model Training:} Finally, we train a student model $f_\phi^{student}$ from scratch using the pseudo-labeled dataset $\hat{\mathcal{D}}_U$:
\begin{equation}
\phi^* = \arg\min_\phi \mathcal{L}(f_\phi^{student}, \hat{\mathcal{D}}_U)
\end{equation}
where $\mathcal{L}$ is the standard supervised loss function used for online HD map construction. After pre-training on pseudo-labels, we fine-tune the student model on the original labeled dataset:
\begin{equation}
\phi^{final} = \arg\min_\phi \mathcal{L}(f_{\phi^*}^{student}, \mathcal{D}_L)
\end{equation}

This two-stage training procedure allows the student model to benefit from both the large-scale pseudo-labeled data and the high-quality ground truth labels.

\section{Experiments}

\subsection{Experimental Setup}

We conduct experiments on the nuScenes dataset~\cite{caesar2020nuscenes} to validate the effectiveness of our teacher-student framework for semi-supervised online HD map construction. We adopt the geographically disjoint split proposed by StreamMapNet~\cite{yuan2024streammapnet}, which ensures no spatial overlap between training and validation sets to prevent data leakage~\cite{lilja2024localization}. To simulate a realistic semi-supervised scenario with limited labeled data, we further split the StreamMapNet training set into two geographically separated subsets: a labeled set $\mathcal{D}_L$ (115 scenes, 4,636 samples, 16.5\%) and a pseudo-unlabeled set $\mathcal{D}_U$ (581 scenes, 23,332 samples, 83.5\%). The pseudo-unlabeled set contains ground truth labels that are used only for evaluation purposes.

Both the teacher and student models are based on UPPM~\cite{peng2025uni}. However, we modify the original UPPM by removing the global map prior component entirely. Our implementation uses only temporal priors, which accumulate and reuse past detection results within each scene. Temporal priors are cleared at scene boundaries, ensuring that each 20-second scene is processed independently without information leakage across scenes. The baseline therefore represents the strongest labeled-data-only configuration of UPPM, ensuring that observed mAP gains from our SSL framework reflect genuine benefit from unlabeled data rather than the effect of enabling temporal modeling.

\subsection{Main Results}

\subsubsection{Teacher Model Training on Labeled Data}

We first train the teacher model based on UPPM from scratch using only the labeled dataset $\mathcal{D}_L$. Table~\ref{tab:main_results} shows the evaluation results on the validation set. This baseline achieved 21.5 mAP. While this is low compared to state-of-the-art methods such as StreamMapNet~\cite{yuan2024streammapnet} and MapDiffusion~\cite{monninger2025mapdiffusion} trained on the full dataset, it is a reasonable result considering that only 16.5\% of the original training set was used for training. This baseline establishes the performance achievable with limited labeled data alone.


\subsubsection{Probability Calibration}

\begin{figure*}[t]
\centering
\includegraphics[width=0.90\textwidth]{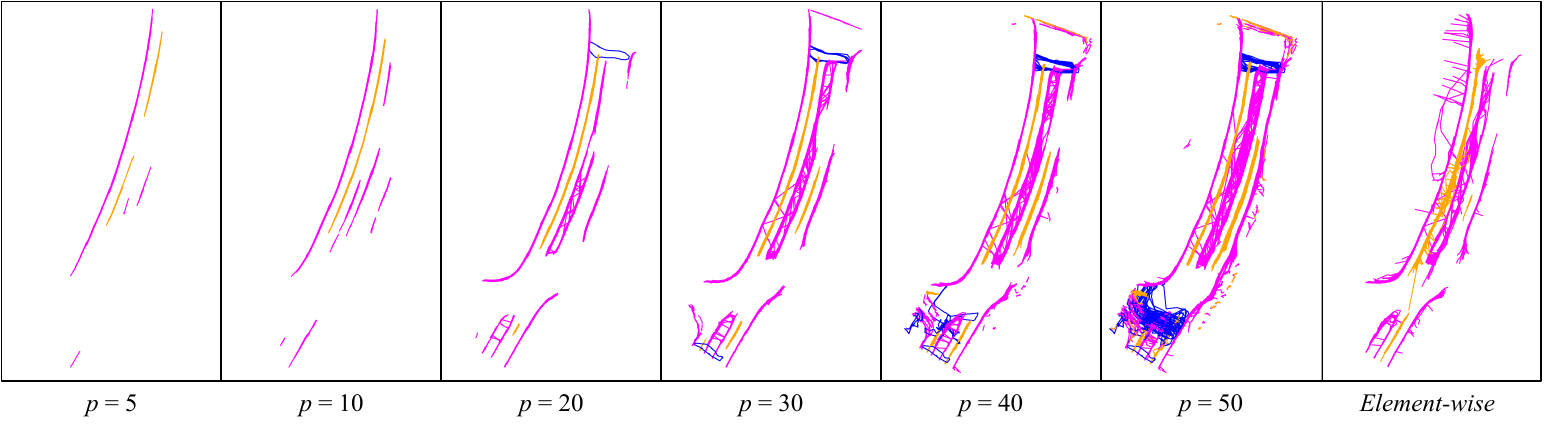}
\caption{Examples of map elements generated by spatial clipping and element-wise filtering. Spatial Clipping (p=5 to p=50) progressively increases coverage at the cost of introducing more uncertain predictions. Element-wise Filtering (threshold=0.3) provides comparable coverage to p=20-30 but cannot preserve partial information from elements with mixed confidence. The visualization demonstrates that Spatial Clipping enables finer-grained control over the precision-coverage trade-off compared to element-level filtering.}
\label{fig:map_priors}
\end{figure*}

\begin{table*}[t]
\centering
    \caption{Effect of percentile threshold on pseudo-label quality. Evaluation on pseudo-unlabeled set using ground truth labels (for evaluation only). ``No Prior'' represents the baseline performance without any map prior, while ``GT Prior'' provides the upper bound of achievable performance with ideal map prior.}
\label{tab:prior_effectiveness}
\begin{tabular}{lcccccccc}
\toprule
Percentile & $AP_{div}$ & $AP_{ped}$ & $AP_{bound}$ & $mAP$ & Dice Mean & Dice Std & IoU Mean & IoU Std \\
\midrule
No Prior & 20.6 & 15.3 & 34.2 & 23.4 & -- & -- & -- & -- \\
$p=5$ & 22.6 & 15.9 & 35.4 & 24.6 & 0.659 & 0.116 & 0.503 & 0.129 \\
$p=10$ & 23.7 & 16.6 & 36.4 & 25.6 & 0.661 & 0.116 & 0.505 & 0.130 \\
$p=20$ & \textbf{24.9} & 16.6 & \textbf{37.8} & \textbf{26.4} & 0.665 & 0.116 & 0.510 & 0.130 \\
$p=30$ & 24.7 & \textbf{16.8} & 37.2 & 26.2 & \textbf{0.668} & 0.116 & \textbf{0.513} & 0.130 \\
$p=40$ & 22.9 & 13.6 & 34.8 & 23.8 & 0.661 & 0.116 & 0.505 & 0.130 \\
$p=50$ & 20.2 & 11.2 & 30.7 & 20.7 & 0.646 & 0.118 & 0.489 & 0.130 \\
GT prior & (35.4) & (25.2) & (47.9) & (36.2) & -- & -- & -- & -- \\
\bottomrule
\end{tabular}
\end{table*}

To ensure reliable confidence estimation, we perform temperature scaling on the teacher model's predictions using the validation set. For calibration, we employ Hungarian matching based on Chamfer distance with a threshold of 1.5m to associate predictions with ground truth, following the MapTRv2 implementation. The optimal temperature parameter is obtained as $T^* = 0.696$. We apply temperature scaling to the classification logits of predicted map elements before sigmoid. Since temperature scaling preserves the relative ordering of confidence scores, using the validation set for calibration does not compromise the fairness of mAP evaluation, which depends only on prediction ranking rather than absolute values.

\subsubsection{Map Prior Generation from Unlabeled Data}

We apply the trained teacher model to the unlabeled dataset $\mathcal{D}_U$ to obtain initial predictions. These predictions are temporally accumulated within each scene and processed through our confidence-aware refinement pipeline to generate map priors. Fig.~\ref{fig:map_priors} visualizes examples of map elements generated by our proposed Spatial Clipping method compared to the baseline Element-wise Filtering approach. For Spatial Clipping, we show results with different percentile thresholds $p \in \{5, 10, 20, 30, 40, 50\}$, demonstrating how the threshold affects the trade-off between precision and coverage. Lower percentiles (e.g., $p=5$) retain only the highest-confidence regions, resulting in sparser but more reliable map priors, while higher percentiles (e.g., $p=50$) provide broader coverage at the cost of including more uncertain predictions.

\subsubsection{Pseudo-Label Generation with Refined Map Priors}

\begin{figure*}[t]
\centering
\includegraphics[width=0.8\textwidth]{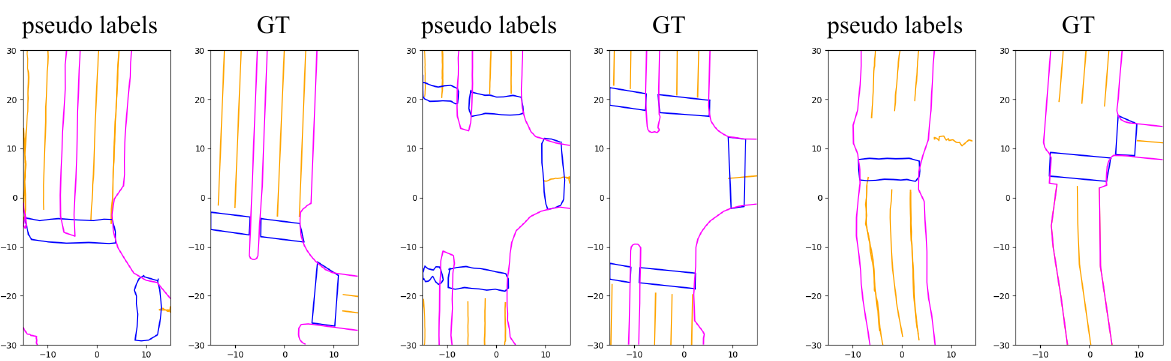}
\caption{Qualitative comparison of pseudo-labels (left) and ground truth (right) for three scenes. The teacher model, conditioned on refined spatial-clipping priors, produces pseudo-labels that closely approximate GT structure despite using only 16.5\% labeled data.}
\label{fig:pseudo_label_vs_gt}
\end{figure*}

We apply the teacher model again to the unlabeled dataset $\mathcal{D}_U$, this time equipped with the refined map priors generated in the previous step. The temporal priors guide the model's attention toward reliable map regions, leading to improved prediction accuracy. To evaluate the effectiveness of different percentile thresholds, we generate pseudo-labels using map priors created with different percentile thresholds $p \in \{5, 10, 20, 30, 40, 50\}$. 
We also include 'No Prior' (p = 0, no filtering) and 'GT Prior' (ground truth map elements) as lower and upper reference bounds respectively, serving as a theoretical performance ceiling for our approach.
Table~\ref{tab:prior_effectiveness} shows the evaluation results in terms of the AP and Dice/IoU metrics on the pseudo-unlabeled set (using ground truth labels available for evaluation purposes only). To compute Dice and IoU, vectorized map elements were rasterized onto 0.5m grids. The results show the mAP peaks at $p=20$, while rasterized Dice/IoU peak at $p=30$ with only a marginal mAP reduction. The mAP difference between p=20 and p=30 is 0.2 points — within noise — while Dice/IoU consistently favor p=30, suggesting the rasterized metric better captures geometric coverage for student training. We therefore choose $p=30$ as a balanced operating point and use it for student training.

\subsubsection{Student Model Training}

Finally, we train the student model from scratch using the pseudo-labeled dataset $\hat{\mathcal{D}}_U$ generated with $p=30$, followed by fine-tuning on the original labeled dataset $\mathcal{D}_L$. Table~\ref{tab:main_results} compares three configurations:

\begin{itemize}
\item \textbf{Baseline:} UPPM trained from scratch on labeled data $\mathcal{D}_L$ only, with temporal priors enabled but conditioned only on the model's own predictions from labeled scenes.
\item \textbf{Ours (Filtering):} Student model trained on pseudo-labels generated using element-wise filtering, then fine-tuned on $\mathcal{D}_L$.
\item \textbf{Ours (Clipping):} Student model trained on pseudo-labels generated using spatial clipping, then fine-tuned on $\mathcal{D}_L$.
\end{itemize}


\begin{table}[t]
\centering
\caption{Semi-supervised learning performance on nuScenes validation set
(StreamMapNet geosplit). Our spatial clipping approach achieves +6.1 mAP
improvement over the UPPM baseline and +2.8 mAP over element filtering.
Results on MapTR (bottom) show the pipeline generalizes to a different
architecture.}
\label{tab:main_results}
\begin{tabular}{llcccc}
\toprule
Backbone & Method & $AP_{div}$ & $AP_{ped}$ & $AP_{bound}$ & mAP \\
\midrule
  UPPM~\cite{peng2025uni} 
 & Baseline           & 21.0 & 5.4 & 38.2 & 21.5 \\
 & Ours (Filtering)   & 26.2 & 6.7 & 41.5 & 24.8 \\
 & Ours (Clipping)    & \textbf{28.7} & \textbf{9.3} & \textbf{44.7} & \textbf{27.6} \\
\midrule
  MapTR~\cite{liao2023maptr}
 & Baseline           & 15.6 & 4.5 & 29.6 & 16.5 \\
 & Ours (Filtering)   & 19.1 & 5.6 & 32.8 & 20.1 \\
 & Ours (Clipping)    & \textbf{20.8} & \textbf{6.9} & \textbf{34.1} & \textbf{21.5} \\
\bottomrule
\end{tabular}
\\[2pt]
\raggedright\footnotesize
†UPPM configurations use temporal priors enabled supported by the
architecture; reported gains reflect SSL contribution only.
\end{table}

The results demonstrate significant performance improvements through our teacher-student framework. Ours (Clipping) achieves a +6.1 mAP improvement over the Baseline, validating the effectiveness of leveraging unlabeled data through confidence-aware pseudo-labeling. Furthermore, Ours (Clipping) outperforms Ours (Filtering) by +2.8 mAP, confirming the superiority of our spatial clipping approach over conventional element-wise filtering. Notably, pedestrian crossing AP improves from 5.4 to 9.3, suggesting that segment-level refinement particularly benefits structurally complex classes. These results validate our hypothesis that preserving partial information from predictions through segment-level refinement leads to more effective utilization of unlabeled data compared to all-or-nothing element-level decisions.


\textbf{Cross-Architecture Generalization} To assess whether our pseudo-labels transfer beyond the UPPM architecture used for the main experiments, we additionally train a MapTR ~\cite{liao2023maptr} student from scratch on the same pseudo-labeled dataset $\mathcal{D}_U$, followed by fine-tuning on $\mathcal{D}_L$, and compare against a MapTR baseline trained on $\mathcal{D}_L$ alone. The MapTR baseline achieves 16.5 mAP, and training with Clipping-based pseudo-labels improves this to 21.5 mAP (+5.0 mAP). The Clipping method also outperforms Filtering method by +1.4 mAP. This result supports our claim that the refinement-and-pseudo-labeling pipeline is model-agnostic: it operates purely on vectorized map elements and does not require MapTR-specific modification.

\section{Conclusion}

We have presented a teacher-student framework for semi-supervised online HD map construction that addresses the critical challenge of labeled data scarcity. Our approach leverages Beta-distribution-based confidence maps to assess the reliability of temporally accumulated predictions, and introduces spatial clipping to selectively preserve high-confidence regions for pseudo-label generation. The refined map elements serve as map priors that improve prediction accuracy in a second pass, resulting in high-quality pseudo-labels for student model training. 
Our baseline uses a fully temporal-prior-enabled UPPM, so the +6.1 mAP gain conservatively reflects genuine SSL benefit. Our framework offers a practical solution to the labeled data scarcity problem, enabling more cost-effective deployment of online HD map construction systems in diverse real-world environments.

\textbf{Limitations.} While our framework demonstrates reasonable improvements, several limitations warrant discussion. First, the temporal accumulation process requires accurate ego-pose estimation and sensor calibration; errors in localization can lead to misaligned map elements and degraded confidence maps. Second, our method assumes that the teacher model's predictions, even if noisy, contain sufficient signal for confidence-based refinement; in scenarios with extremely limited labeled data or severe domain shift, the initial teacher quality may be insufficient. Additionally, an $\approx$10 mAP gap remains between our refined pseudo-labels and the GT Prior upper bound (Table 1), indicating headroom attributable to limited teacher quality at 16.5\% labeled data. Finally, we evaluate only on nuScenes; extending to additional datasets and cross-region settings is future work.


%
%
\bibliographystyle{splncs04}
\bibliography{main}
\end{document}